\documentclass[letterpaper]{article} 
\usepackage{aaai23}  
\usepackage{times}  
\usepackage{helvet}  
\usepackage{courier}  
\usepackage[hyphens]{url}  
\usepackage{graphicx} 
\usepackage{natbib}  
\usepackage{caption} 
\usepackage{algorithm}
\usepackage{tikz}
\usepackage{subcaption}
\usepackage{array}
\usepackage{amsmath}    
\usepackage{cuted}
\usepackage[spaceRequire=false]{algpseudocodex}
\usepackage{newfloat}
\usepackage{listings}
\DeclareCaptionStyle{ruled}{labelfont=normalfont,labelsep=colon,strut=off} 
\floatstyle{ruled}
\newfloat{listing}{tb}{lst}{}
\floatname{listing}{Listing}
\title{Assessing Systematic Weaknesses of DNNs using Counterfactuals}
\author {
    Sujan Sai Gannamaneni\textsuperscript{\rm 1},
    Michael Mock\textsuperscript{\rm 1}, 
    Maram Akila\textsuperscript{\rm 1}
}
\affiliations {
    \textsuperscript{\rm 1} Fraunhofer IAIS\\
    \{sujan.sai.gannamaneni, michael.mock, maram.akila\}@iais.fraunhofer.de
}

\newcommand{\Xx}{\mathcal{X}}
\newcommand{\Yy}{\mathcal{Y}}
\newcommand{\Dd}{\mathcal{D}}
\newcommand{\Cc}{\mathcal{C}}
\newcommand{\Ss}{\mathcal{S}}
\newcommand{\disc}{\Delta}
\newcommand{\diff}{\mathcal{A}}
\newcommand{\conv}{\text{con}}
\newcommand{\rnd}{\text{rnd}}
\newcommand{\cf}{\text{cf}}
\newcommand{\res}{\text{res}}
\newcommand{\pair}{\text{pair}}
\newcommand{\expect}{\mathsf{E}}
\newcommand{\cfi}[1]{{\operatorname{CFI}{\left( #1 \right)}}}

\DeclareMathOperator*{\argmin}{arg\,min}
\newcommand{\semdim}{d}
\newcommand{\semdist}{\operatorname{dist}_\semdim}
\newcommand{\dist}{\texttt{dist}}
\newcommand{\numpix}{\texttt{num\_pixels}}
\newcommand{\visibility}{\texttt{visibility}}
\newcommand{\assid}{\texttt{asset\_id}}
\newcommand{\size}{\texttt{size}}
\newcommand{\sd}{s}
\newcommand{\snum}{n}
\newcolumntype{P}[1]{>{\centering\arraybackslash}p{#1}}

\begin{document}

\maketitle

\begin{abstract}
With the advancement of DNNs into safety-critical applications, testing approaches for such models have gained more attention.
A current direction is the search for and identification of systematic weaknesses that put safety assumptions based on average performance values at risk.
Such weaknesses can take on the form of (semantically coherent) subsets or areas in the input space where a DNN performs systematically worse than its expected average.
However, it is non-trivial to attribute the reason for such observed low performances to the specific semantic features that describe the subset.
For instance, inhomogeneities within the data w.r.t.\ other (non-considered) attributes might distort results.
However, taking into account all (available) attributes and their interaction is often computationally highly expensive.
Inspired by counterfactual explanations, we propose an effective and computationally cheap algorithm to validate the semantic attribution of existing subsets, i.e., to check whether the identified attribute is likely to have caused the degraded performance. We demonstrate this approach on an example from the autonomous driving domain using highly annotated simulated data, where we show for a semantic segmentation model that (i) performance differences among the different pedestrian assets exist, but (ii) only in some cases is the asset type itself the reason for this reduction in the performance.

\end{abstract}

\section{Introduction}
\label{section:introduction}

Recently, there has been great interest in deploying deep neural networks (DNNs) for computer vision tasks in safety-critical applications like autonomous driving~\cite{siam2018comparative} or medical diagnostics~\cite{hesamian2019deep}. However, rigorous testing for verification and validation (V\&V) of DNNs is still an open problem. Without sufficient V\&V, using DNNs for such safety-critical applications can lead to dangerous situations.
While a large body of work has focused on improving robustness~\cite{hendrycks2019augmix}, defending against adversarial attacks~\cite{akhtar2018threat, chakraborty2018adversarial}, and improving domain generalization~\cite{oza2021unsupervised, wang2018deep}, only recently, a few works have focused on identifying systematic weaknesses of DNNs~\cite{d2022spotlight, eyuboglu2022domino, Gannamaneni_2021_ICCV, lyssenko2021evaluation, syed2020dnn}. Such systematic weaknesses, however, pose a significant challenge from a safety perspective, as seen by several real-world examples~\cite{buolamwini2018gender, de2019does}, where DNNs have been shown to perform worse for certain subsets of data. 
Such effects have been extensively studied in the context of Fairness, see, e.g., \cite{wang2020towards}, where often bias due to skin color or gender is addressed.
However, this is not sufficient to investigate the safety of a DNN. Instead, all dimensions that could potentially influence performance have to be taken into account.

This can be seen in the broader context of trustworthiness assessments for ML, specifically regarding reliability.
Multiple high-level requirements or upcoming standards point out the issues of data completeness or coverage to varying levels of detail.
For instance, \cite{hleg} discusses both data completeness in the context of Fairness (as presented above) but also states that the robustness of the application is bound to a specific application domain.
Other approaches, such as \cite{VDE_SPEC_90012}, see a rigorous definition of the application's input space and aligning the used data to it as an important aspect of trustworthiness.
Likely, such approaches will reside on semantic, human-understandable definitions of data dimensions to define the domain and, potentially, also to demonstrate coverage.

An informal example to illustrate the concept of data specification and its relation to potential systematic weaknesses could be a classifier that distinguishes between cats and dogs.
While it would be a common procedure to investigate the performance of such a classifier by, e.g., a confusion matrix between the two classes, one could (and for critical algorithms) should extend both the task description and the investigation to include sub-classes.
For instance, one could evaluate the classifier's performance on the various dog breeds the classifier was built for (definition of the input space) and ascertain whether each of these breeds is recognized with sufficient performance.
To achieve the latter, data, in most cases, must come with sufficiently detailed attribution.
Obtaining such meta-information for unstructured data, e.g., images as used for object detection is a challenging objective in its own right.
Recent works, see, for instance, \cite{d2022spotlight, eyuboglu2022domino}, use dedicated neural networks or parts of the investigated networks to obtain information on the data while other approaches, such as ours~\cite{Gannamaneni_2021_ICCV} or, e.g.,~\cite{fingscheidt2022deep}, focus on a controlled data generation process.
For example, by using synthetic data, which yields detailed data description as a ``by-product''.
The latter, more labeling-oriented approaches, have the advantage that obtained information is aligned with the semantic dimensions of the data.
For the former approaches, outcomes are often raw data subsets whose descriptive attributes have to be uncovered in an additional step.

The analysis of identified (potential) weaknesses is often more complex as multiple data dimensions can influence the outcome simultaneously.
For instance, in the above example, is the dog breed a performance limiting factor, or can the weakness be attributed to another factor, either in a causal (e.g.,\ size of the breed) or only correlated fashion (e.g.,\ if many images of one dog breed were taken by a different camera model)?
Answering such questions requires a detailed (meta-)description of the data, which is challenging outside of synthetic data and requires consideration of the impact of multiple dimensions and their interactions.
The latter point often leads to a ``combinatorial explosion'' as soon as two, and higher order interactions are taken into account.
At the same time, correct identification of data subsets with weak performance is important as it is otherwise hard to mitigate such systematic failure modes.
For instance, the common approach for mitigation is by generating or obtaining more data samples and adding them to the training data.

As data gathering or generation is costly, it is important to understand better which data is needed to improve coverage and training.
Towards this goal, we propose a computationally cheap method that helps to determine whether identified performance differences between two semantically distinguished subsets $\Xx$ and $\Yy$ can be attributed to either the semantically distinguishing property between the sets or other described but not considered factors.
For this, we adapt the concept of counterfactual explanations~\cite{verma2020counterfactual}, where we pair the elements of $\Xx$ only to those elements in $\Yy$, which are most similar w.r.t. all other known attributes.
Suppose our method determines that the distinguishing semantic property between the sets can be considered as the true cause for the performance difference. In that case, generating more data for the weak set should be a useful measure to improve the DNN performance.
If not, the method allows for identifying semantic subsets of $\Xx$ or $\Yy$, respectively, that are likely to contain the true cause of the performance gap.

\section{Related work}
\label{section:related_work}

In this section, we discuss literature related to finding systematic weaknesses of machine learning models by classifying them based on the input data type. We first present approaches working with structured, e.g.,\ tabular data, and then present approaches for unstructured data such as images.
Drawbacks of the existing approaches are presented subsequently.
Lastly, we introduce works on counterfactuals.

For structured data, SliceFinder~\cite{chung2019slice} identifies weak slices (subsets) of data by ordering slices based on certain criteria like performance, effect, and slice size.
These weak slices correspond to the systematic weakness of the models.
Similarly, SliceLine~\cite{sagadeeva2021sliceline} provides an enumeration of all slice combinations using a scoring function and different pruning methods.
In addition, subgroup search~\cite{atzmueller2015subgroup, herrera2011overview} is an extensively researched field in data mining for identifying interesting subgroups or subsets of data based on certain quality criteria.
However, at higher dimensions, subgroup search methods are computationally very intensive.
Complexity grows even further when going from structured to unstructured data, such as images.
In these cases, spaces are intrinsically (very) high dimensional.
Accordingly, there has been no widespread use of such methods for computer vision.

Due to this complexity and the difficulty of obtaining metadata for unstructured data like images, some approaches have been proposed to use intrinsic information from the DNNs.
\citet{eyuboglu2022domino} developed DOMINO for finding systematic weaknesses by using cross-modal representations generated from a pre-trained CLIP~\cite{radford2021learning} model.
However, using CLIP, a black-box DNN model in the embedding generation process, adds further complexity to validating the DNN-under-test.
Spotlight~\cite{d2022spotlight} uses representations from the final layers of DNNs to identify contiguous regions of high loss.
These contiguous regions (slices) with the highest loss are then considered weak slices that could lead to potential systematic weaknesses of the DNN.
Both these approaches are restricted to classification tasks.
Furthermore, due to the nature of the approach, resulting weaknesses need to be interpreted, either by other ML models or humans, to derive actionable insight into its nature, i.e., to identify the common cause of all found weakly performing elements.
This approach hinders a systematic evaluation and might also be prone to errors.

With a particular focus on autonomous driving, several works~\cite{Gannamaneni_2021_ICCV,lyssenko2021evaluation,syed2020dnn} have used computer simulators to generate metadata to identify systematic weaknesses in object detection and semantic segmentation models.
In our earlier work~\cite{Gannamaneni_2021_ICCV}, we made modifications to the Carla simulator~\cite{dosovitskiy2017carla} to generate pedestrian-level metadata. 
With the generated data, we trained a DeepLabv3+~\cite{chen2017deeplab} model and identified several (potential) systematic weaknesses w.r.t.\ digital asset type and skin color.
Similarly,~\citet{lyssenko2021evaluation} evaluated DNN performance along a semantic feature, the distance of the pedestrian to the ego-vehicle, using data generated from their modified Carla simulator and showed a linear decrease in performance with increasing distance.
\citet{syed2020dnn} proposed a validation engine, `VALERIE', and evaluated the performance of two different DNNs w.r.t.\ metadata attributes like pixel-area, occlusion-rate, and distance of pedestrians.
All those works circumvent complexity by investigating features in isolation. In this work, we, however, take all (available) features into account.

With regards to counterfactuals, \citet{verma2020counterfactual} show that a large body of work in explainable AI has proposed counterfactual methods to provide explanations of DNN behavior by investigating ``what if'' scenarios.
Several methods~\cite{dandl2020multi, goyal2019counterfactual, kanamori2020dace, mothilal2020explaining, wachter2017counterfactual} structure finding counterfactuals as an optimization problem using gradients of the model-under-test similar to finding adversarial attacks.
However, unlike in adversarial attacks, counterfactual methods can possess several additional constraints~\cite{verma2020counterfactual} such as validity~\cite{wachter2017counterfactual}, actionability~\cite{karimi2021algorithmic, ustun2019actionable}, closeness to data manifold~\cite{dhurandhar2018explanations, joshi2019towards, mahajan2019preserving} and/or sparsity in feature changes~\cite{guidotti2018local, karimi2020model}.
Counterfactuals also differ from feature attribution methods like LIME~\cite{ribeiro2016should} or Shapley Values~\cite{lundberg2017unified} as counterfactuals identify new inputs which lead to change in predictions rather than attributing predictions to a set of features. 

While these methods have mostly restricted themselves to tabular data and simple image datasets like MNIST~\cite{lecun2010mnist}, our method performs counterfactual evaluation on a semantic segmentation dataset for autonomous driving using metadata descriptions of pedestrians in the images.
In addition, the goal of most counterfactual methods, as they concern fairness, is to (actively) change a decision or output, while our focus is on investigating performance differences.
Furthermore, the gradient-based approaches require additional inference steps to investigate their (newly created) counterfactual sample.
Our method instead uses a statistical formulation of counterfactuals~\cite{pearl2019seven} as for unstructured data creating new ``what if'' scenarios is (computationally) too expensive, even when using simulators, to afford a meaningful analysis over many data points.


\section{Method}
\label{section:method}

In this section, we provide a general description of our counterfactual algorithm and how it may be used to investigate performance differences within a given dataset $\Dd$.
As our approach is general, we relegate instantiating $\Dd$ (and its properties) to the next section.
Here, it is sufficient that all elements of $\Dd$ can be seen as individual inputs or points of interest (later: those will be separate pedestrians, and the task will be their recognition), for each of which we can obtain a respective performance value (later: intersection-over-union (IoU)) and a semantic description based on multiple dimensions $\Ss = \{\sd_1,\ldots,\sd_\snum\}$ (later: for instance, distance from the vehicle or asset membership).

As motivated, we are interested in identifying DNN weaknesses in terms of data subsets that perform weakly due to properties inherent to the data, e.g.,\ occlusion or under-represented input types.
Furthermore, we want to validate whether the decreased performance can be attributed to the identified property.

For this, let us consider slices $\Xx,\Yy \subseteq \Dd$ of the data where variances in performance exist between them.
Such slices can be obtained, e.g., by selecting all elements which have one (or multiple) fixed semantic properties (or ranges thereof) in common,
e.g.,\ $\Xx = \{ x \, | \, x\in \Dd \wedge \sd_\text{asset}(x) = \text{asset}_1 \}$.
For this notation, we assume that $P(x)$ and $P(\Xx)$ are the performance value of the element $x\in\Xx$ or the set of performance values over all of $\Xx$, respectively.\footnote{These statements hold for all subsets of $\Dd$, e.g., also for elements of $\Yy$.}
This allows us to write the ``conventional'' performance difference between the two slices as
\begin{equation}
    \disc_\conv (\Xx,\Yy) = \mu[P(\Xx)] - \mu[P(\Yy)] \,
    \label{eq:conv}
\end{equation}
where $\mu$ denotes the mean-value of the respective set.
We may also formulate a related quantity based on the set of local performance differences
\begin{equation}         
\begin{split}
    \diff_\rnd(\Xx,\Yy) = \left\{P(x_i) - P(y_i) \quad|\quad \right. \\ 
    \left. x_i \in \Xx, y_i \in \Yy, i\in \{1,\ldots,\min(|\Xx|,|\Yy|)\} \right\}
    \,,
    \label{eq:diffRnd}
\end{split}
\end{equation}
where $i$ provides a fixed but arbitrary index of the sets $\Xx,\Yy$ respectively.
If we were to average over all possible such indices, we would find that
\begin{equation}
    \disc_\conv (\Xx,\Yy) = \expect[\disc_\rnd (\Xx,\Yy)]
    \label{eq:similarityConvRnd}
\end{equation}
holds in expectation with $\disc_\rnd (\Xx,\Yy) = \mu[\diff_\rnd (\Xx,\Yy)]$.\footnote{This is the case as the average over means of random subsets converges to the mean of the full set. In the special case of $|\Xx| = |\Yy|$ we find $\disc_\conv (\Xx,\Yy) = \disc_\rnd (\Xx,\Yy)$ directly.}
Looking at eq.\ (\ref{eq:diffRnd}), we observe that differences disregard the semantic contexts of the pairings $x_i,y_i$ and could compare datapoints with entirely different properties (e.g.,\ pedestrians of high occlusion with clearly visible ones).
We, therefore, build a dedicated paired dataset $\Cc_\cf (\Xx, \Yy)$, where for each element of the reference dataset $\Xx$ we find the most similar (by semantic description) datapoint in the other dataset $\Yy$ as shown in equation~(\ref{eq:eq4}),
\begin{figure*}[ht]
\begin{equation}
    \Cc_\cf (\Xx, \Yy) = \left\{
    \begin{array}{ll}
        \{ (x_i, y_\cfi{x_i,\Yy}) \quad|\quad i\in\{1,\ldots, |\Xx|\} \} &  \quad\text{if } |\Xx|\leq |\Yy|\\
        \{ (x_\cfi{y_i,\Xx}, y_i) \quad|\quad i\in\{1,\ldots, |\Yy|\} \} &  \quad\text{if } |\Xx|> |\Yy|
    \end{array}
    \right.
\label{eq:eq4}   
\end{equation}
\end{figure*}
where the counterfactual datapoints in $\Yy$ are selected by
\begin{equation}
\begin{split}
    \cfi{x_i,\Yy} = \argmin_j \semdist(x_i,y_j)\quad\\\text{where } y_j\in \Yy \backslash \{x_i\} \,
    \label{eq:cfi}
\end{split}
\end{equation}
which minimizes the distance $\semdist$ using a subset $\semdim \subseteq \Ss$ of the semantic features $\Ss$.\footnote{In cases where the $\argmin$ is not unique, we select the first element given a fixed but arbitrary order of all elements in $\Dd$. Ideally, $\semdim$ should exclude those elements of $\Ss$ that were used to construct $\Xx,\Yy \subseteq \Dd$.}
Refer to the next section for a concrete example of our metric. 
Based on this paired set we can calculate the set of counterfactual differences
\begin{equation}
\begin{split}
    \diff_\cf^\tau (\Xx,\Yy) = \{ P(x_i) - P(y_i) \quad|\quad \\ (x_i,y_i) \in \Cc_\cf (\Xx, \Yy) \wedge \semdist(x_i,y_j) \leq \tau \}\,
    \label{eq:acf}
\end{split}
\end{equation}
taking into account only those pairings that are closer than a threshold $\tau$ to ensure sufficiently close proximity of points.
The corresponding average performance difference is given by $\disc_\cf(\Xx,\Yy) = \mu[\diff_\cf^\tau (\Xx,\Yy)]$.
Please note that, importantly, the counterfactual difference $\disc_\cf$ does not have to coincide with $\disc_\conv$.
If, for instance, $| \disc_\cf | \ll |\disc_\conv |$ the performance differences between $\Xx$ and $\Yy$ can likely not be attributed to the semantics used to separate both sets from $\Dd$, but instead are a property of some other latent discrepancy between the two sets.
In these cases, one might want to investigate the non-matched elements, i.e.,\ those elements of $\Xx,\Yy$ that are not part of $\Cc_\cf$, as they might carry another distinguishing attribute.
Please note that the matching procedure of eq.\ (\ref{eq:cfi}) is, in most cases, not commutative between the sets.
However, when $\Xx,\Yy$ do not intersect our fixed order of operations ensures $\disc_\cf(\Xx,\Yy)=- \disc_\cf(\Yy,\Xx)$.
Conversely, if $\Xx=\Yy$, our definition avoids collapse as the points cannot be matched onto themselves.
A more procedural definition to determine the counterfactual difference $\disc_\cf$ can be found in algorithm \ref{alg:counterfactual} below.
\begin{algorithm}[thb]
\caption{Computation of the counterfactual performance difference $\disc_\cf(\Xx, \Yy)$}
\label{alg:counterfactual}
\begin{algorithmic}
    \LComment{Sets w.r.t those the difference is calculated:}
    \Require $\Xx, \Yy$
    \LComment{Hyper-parameters for distance calculation:}
    \Require $\tau, \semdim$
    \State $c \gets 0$
    \Comment{Counter variable}
    \State $\disc_\cf \gets 0$
    \Comment{Result variable}
    \LComment{Iterate over all entries of the smaller set:}
    \For{$i=1,\ldots,\min\left(|\Xx|,|\Yy|\right)$}
        \LComment{Use smaller set as reference:}
        \If{$|\Xx| \leq |\Yy|$}
            \State $j \gets \cfi{x_i,\Yy}$
            \Comment{See eq. (\ref{eq:cfi}), $x_i\in \Xx$}
            \State $i' \gets i$
        \Else
            \State $j \gets \cfi{y_i,\Xx}$
            \State $i' \gets j$
            \Comment{Exchange index values via $i'$}
            \State $j \gets i$
        \EndIf
        \State $\delta \gets \semdist(x_{i'},y_j)$
        \LComment{Only use result if points are near one another:}
        \If{$\delta\leq\tau$}
            \State $\disc_\cf \gets \disc_\cf + \left(P(x_{i'})-P(y_j)\right)$
            \Comment{$P(x)$: performance of $x$}
            \State $c = c +1$
            \Comment{Increment counter}
        \EndIf
    \EndFor
    \If{$c>0$}
        \State 
        \Return $\disc_\cf / c$
        \Comment{Yields mean perf. difference}
    \Else
        \State 
        \Return Null
        \Comment{Sets could not be matched for selected max.\ distance of $\tau$}
    \EndIf
\end{algorithmic}
\end{algorithm}


\section{Experimental Setup}
\label{section:experiment_setup}

In this section, we describe the used dataset, the DNN-under-test and the concrete implementation of the counterfactual calculations and their metrics.

\textbf{Dataset} \textendash\
For our experiments, we use the dataset generated from our previous work,~\citet{Gannamaneni_2021_ICCV}, which contains extracted meta-information\footnote{https://github.com/sujan-sai-g/Semantic-Concept-Testing-using-CARLA} for each pedestrian visible within an image.
It was generated using Carla Simulator v0.9.11~\cite{dosovitskiy2017carla} and contains 23 classes following a mapping similar to Cityscapes~\cite{cordts2016cityscapes}.
It consists of images with traffic scenes, corresponding semantic segmentation ground truth, and pedestrian meta-information.
This way, and using additional computer vision post-processing, we obtain the set $\Ss$ of our semantic features encompassing \{\dist, \visibility, \numpix, \size, \assid, $x_\text{min}, y_\text{min}, x_\text{max}, y_\text{max}$\}.
Here, \dist\ refers to the euclidean distance of the pedestrian from the ego vehicle, \visibility\ to the percentage of the pedestrian that is unoccluded, \numpix\ to the (absolute) number of pixels belonging to the pedestrian, \assid\ gives an identifier for the 3D model used by the simulator.
The coordinates $x,y$ belong to the bounding box of the pedestrian, and \size\ provides its respective area.
With this setup, we generated $7\,394$ images (all from the ``Town02'' map in Carla) and trained the semantic segmentation model DeepLabv3+~\cite{chen2017deeplab} on them. We follow the same training setup and data pre-processing as our prior work~\cite{Gannamaneni_2021_ICCV}.
To investigate the performance of the said model, we added the individually achieved IoU (Intersection-over-Union) to the collection of our per-pedestrian metadata.
In the remainder, all analysis will be performed on this resulting table of performance and meta-data descriptions containing a total of $24\,424$ entries.\footnote{The full table contains even more pedestrians, we, however, limit ourselves to true positives ($IoU > 0$), exclude entities beyond a euclidean distance threshold of $100$ (as their detection is rather a chance event and not indicative of actual DNN performance) and removed bounding-boxes larger than $10^5$ pixels (as they constitute rare outliers).}

\textbf{Implementation of Counterfactual Similarity}\ \textendash\ 

To build counterfactual datasets as given by eq.~(\ref{eq:eq4}), we need to specify a distance among the semantic dimensions.
For this, we use the euclidean metric, where we used one-hot-encoding for the categorical \assid\ and re-scaled all numerical dimensions to the unit range $[0,1]$.
If not stated otherwise, the cut-off parameter $\tau$, see eq.~(\ref{eq:acf}), is chosen as 0.2.
In the cases where $\Xx$ and $\Yy$ do not intersect, we can calculate counterfactual pairings (more) efficiently using a ball-tree algorithm to find the nearest points.
More concretely, the problem can be seen as a $k$-NN classification for $k=1$, where we are interested only in the performance of the nearest point in $\Yy$ w.r.t.\ a sample from $\Xx$.


\section{Results}
\label{section:results}

We conducted three different experiments.
First, we evaluate the expressive power of semantic features in the similarity search.
These results provide insight into the noise level of the performance values and to which degree they depend on the known semantic attributes.
In the second subsection, the results of the counterfactual analysis for a semantic-dimension-under-test are provided and discussed.
Finally, using results from the counterfactual analysis, we show that interesting subsets of the slices can be discovered, which can be used by the developer to further narrow down which data dimensions to investigate.

\subsection{Evaluating expressive power of semantic features}
\begin{figure*}[!htb]
     \centering
     \begin{subfigure}[b]{0.43\textwidth}
         \centering
        \includegraphics[width=\textwidth]{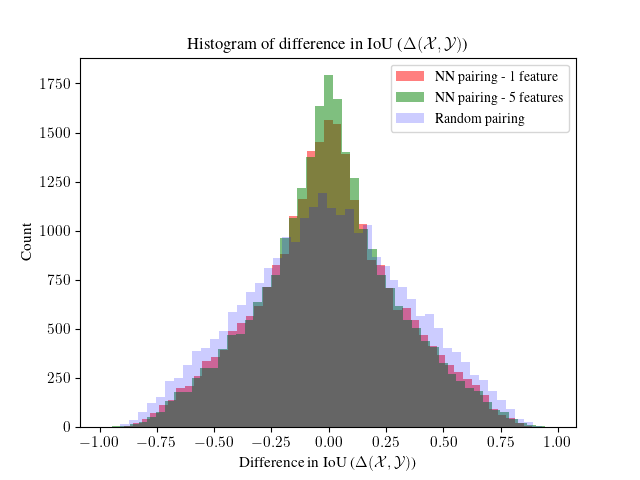}
     \end{subfigure}
     \hfill
     \begin{subfigure}[b]{0.43\textwidth}
         \centering
        \includegraphics[width=\textwidth]{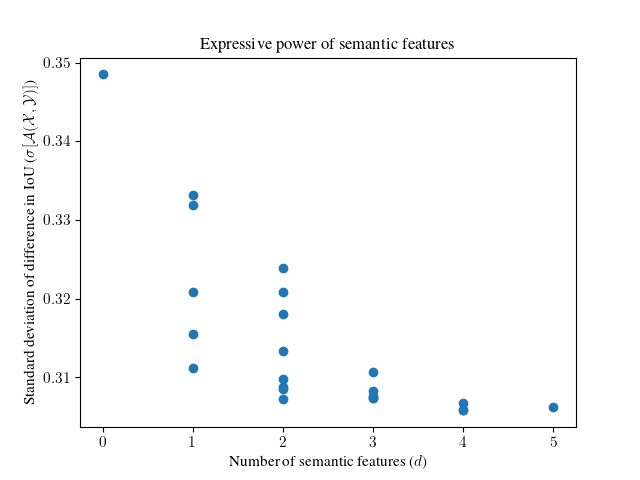}
     \end{subfigure}
     \hfill
     \caption{Left: The histogram depicting the difference in performance for samples between two datasets using three matching techniques: using a single feature with nearest neighbor matching (red), using five features with nearest neighbor matching (green), and random matching (blue), (Best seen in color). Right: The reduction in the standard deviation of the performance difference distributions as more semantic features are used in the NN search.}
     \label{fig:sanity_check}
\end{figure*}

Our counterfactual analysis assumes that the semantic features are expressive enough to provide \textit{on average} some indication on the performance of the DNN-under-test, i.e., that more similar points are more likely to have a similar performance.
We perform the following check to validate this assumption:
Setting $\Xx=\Yy=\Dd$, we can evaluate the statistics of $\diff_\cf^\infty(\Dd,\Dd)$, see eq.~(\ref{eq:acf}), while including an increasing number of semantic dimensions into $\semdim$, the special case of $\semdim=\{\}$ corresponds to a purely random pairing of the elements.
To provide a baseline, assuming a flat i.i.d.\ distribution of the performance values in the full range of $[0,1]$ $\diff_\cf$ would follow a triangular distribution with element-wise differences ranging from $-1$ to $+1$.
The standard deviation of such a triangular distribution would be $\sigma_\text{tri}=1/\sqrt{6}\approx 0.41$.
Any decrease of $\sigma[\diff_\cf^\infty]$ from this threshold indicates a deviation from pure randomness.
We provide an overview of the development of $\sigma$ on the r.h.s.\ of Figure~\ref{fig:sanity_check} for increasing numbers of used semantic properties.
The random pairing ($|\semdim|=0$) is still close to the threshold with $\sigma[\diff_\rnd]\approx 0.35$.
Taking a single feature into account ($|\semdim| = 1$) already leads to a decrease of (depending on the feature selected) up to $13\%$, while including multiple features leads to stronger but saturating decay.\footnote{Here, for $|d|=5$ we use all features of $\Ss$ except for the categorical \assid\ and limit the bounding box coordinates to $x_\text{min}$.}
The l.h.s.\ of Figure~\ref{fig:sanity_check} provides a graphical representation of this statement by showing histograms of $\diff_\cf^\infty$  for some selected $\semdim$.

We can interpret the decay of the standard deviation further by comparison to a toy experiment.
For this, we consider two uniform distributions, each with standard deviation $\sigma_\text{sing}$, that are shifted against one another by an offset $\Delta\mu$.
We can see the membership of an element to either of these distributions as a semantic property of this toy example.
If we are unaware of it and investigate elements that are equally likely to be drawn from either of the distributions, we will naturally observe a larger spread of $\sigma_\text{both}$.
If we identify this spread with the standard deviation we have seen for the case of $|d|=0$ and likewise use the $|d|=5$ case as value for $\sigma_\text{sing}$, we can estimate (in a rough fashion) the scale of potential shifts $\Delta\mu$. For this, we use
\begin{equation}
\Delta \mu = \sqrt{2\left(\sigma_\text{both}^2-\sigma_\text{sing}^2 \right)}\,,
\label{eq:DeltaMu}
\end{equation}
which holds for the toy model only.
It, however, suggests a value of $\Delta\mu\approx 0.2$ as an approximate range.

\subsection{Counterfactual analysis}

\begin{figure*}[!htb]
     \centering
     \begin{subfigure}[b]{0.45\textwidth}
         \centering
         \includegraphics[width=\textwidth]{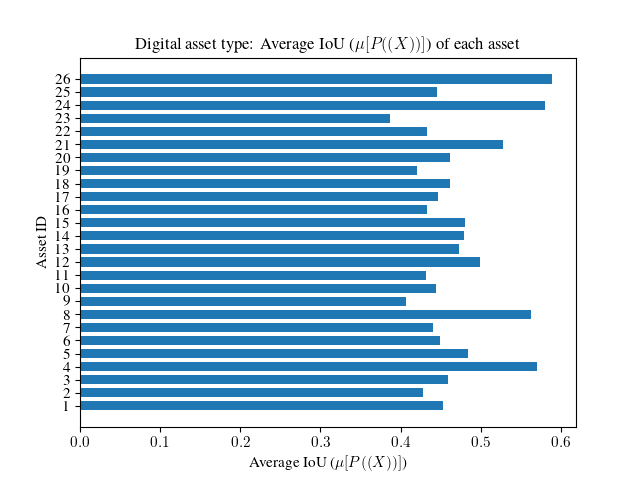}
     \end{subfigure}
     \hfill
     \begin{subfigure}[b]{0.45\textwidth}
         \centering
         \includegraphics[width=\textwidth]{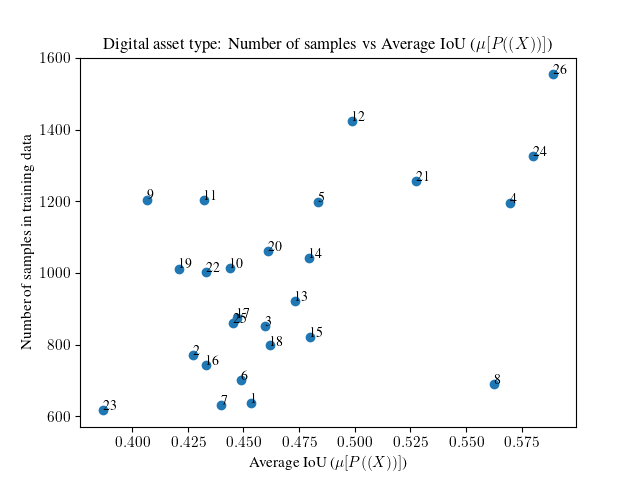}
        
     \end{subfigure}
     \hfill
     
     \caption{Left: Comparison of the average performance of the different digital assets. The numbers on the y-axis represent the unique asset IDs. Right: Comparison of the average performance of the different assets and the count of samples in the training data. The numbers next to a point represent the asset ID.}
     \label{fig:count_vs_iou}
\end{figure*}

Having established that using all five semantic features is beneficial in the similarity search, we focus on the left-out \assid\ to identify if there are indeed systematic weaknesses present for slices in this semantic feature (semantic-feature-under-test).
To identify interesting $(\Xx,\Yy)$ combinations, we make use of the average performance of the assets for the entire training data as shown on the left in Figure~\ref{fig:count_vs_iou}.
We consider the highest performing assets (26, 24, 4, 8) and the lowest performing assets (9, 23) as candidates for counterfactual analysis.
The results for different combinations of these assets are shown in Table~\ref{tab:counterfactual_results}.
For each combination, we provide the conventional difference in performance, $\disc_\conv$ see eq.\ (\ref{eq:conv}), the counterfactual difference in performance, $\disc_\cf$ eq.\ (\ref{eq:cfi}), and the random pairing difference, $\disc_\rnd$ eq.\ (\ref{eq:diffRnd}).
The latter we provide as a sanity check and find that it, as expected according to eq.~(\ref{eq:similarityConvRnd}), approximately follows the conventional difference.\footnote{One could expect deviations in this quantity, if either the sample size becomes too small or there are strong inhomogeneities in the distribution of the data.}
Additionally, observed performance differences do roughly abide, in their maximal values, to the scale of $\Delta\mu$ from the previous section.
However, when comparing counterfactual and conventional performance differences, we, in some cases find strong discrepancies.
In the case of the weakly performing asset 9, the counterfactual difference to all strong performing assets is negligible, suggesting that the decreased performance of this asset is due to the presence of pedestrians that are challenging due to some properties other than their membership to asset 9, i.e., this asset (in itself) does not constitute a systematic weakness.
Hence, just generating more (training) data for asset 9 would be an ineffective way to increase its performance.
Yet, the circumstances leading to the decreased performance in the slice, even if unrelated to the asset type, should be investigated further; see the results from the next experiment. 
In contrast to asset 9, for asset 23 (the other weak candidate), the counterfactual differences rather emphasize the performance discrepancy to all well-performing assets making it a likely candidate for an actual systematic weakness.
Therefore, generating more data for asset 23 could be useful.
This is also supported by investigating the number of samples of the different assets in the training data and the average performance per asset as shown on the right in Figure~\ref{fig:count_vs_iou}.
We can see that assets 4, 9, and 24 have a similar amount of samples.
However, 23 only has half the numbers compared to 26 (the asset with the highest performance).

\begin{table*}[htb]
 \caption{\label{tab:counterfactual_results} For different $\Xx$, $\Yy$ combinations, the difference in mean performance using the conventional, random, and nearest neighbor pairing is shown. For convenience, $\Yy$ denotes the weaker performing set.}
\centering
\begin{tabular}{ccccc}
 Asset slices $(\Xx,\Yy)$ & $\disc_\conv (\Xx,\Yy)$ & $\disc_\cf (\Xx,\Yy)$ & $\disc_\rnd (\Xx,\Yy)$ & Result \\
 \hline
 4, 9   & 0.16306 &  0.00887    & 0.15294 &  $\disc_\conv \neq \disc_\cf$  \\
  8, 9  & 0.15588 &   0.05064    & 0.15281  &  $\disc_\conv \neq \disc_\cf$\\
 24, 9 & 0.17350 & 0.02292 & 0.16416 & $\disc_\conv \neq \disc_\cf$ \\
 
 26, 9 & 0.18270 & -0.02688 & 0.18341 & $\disc_\conv \neq \disc_\cf$ \\
 4, 23  & 0.18303 &  0.22153    & 0.17468  & $\disc_\conv \sim \disc_\cf$ \\
  8, 23 &  0.17585 & 0.23775 & 0.17626 & $\disc_\conv \sim \disc_\cf$ \\
  24, 23 & 0.19347 & 0.22053 & 0.18952 & $\disc_\conv \sim \disc_\cf$ \\

 26, 23 & 0.20267 &  0.23950    & 0.19659 &  $\disc_\conv \sim \disc_\cf$  \\
 \hline

\end{tabular}
\end{table*}

\subsection{Discovering residual subsets}

As discussed above, asset 9 is underperforming, but asset membership does not seem to constitute the true cause of the issue.
Hence, just generating more data for this asset is unlikely to resolve the problem efficiently, which is also supported by the fact that there is already a relatively high number of data available for asset 9 (see right-hand side of Figure~\ref{fig:count_vs_iou}).
For example, looking at asset 8, we see that it has much less data, but significantly higher performance than asset 9.
In the following, we intend to identify, purely based on the semantic description, the data causing this effect.

When building counterfactual pairs among the samples from $\Xx$, $\Yy$, only a subset of the larger set (here w.l.o.g.\ named) $\Yy$ might be used.

In such a case, this provides a way to slice $\Yy$ into two subsets, i.e., a paired subset
\begin{equation}
    \Yy_\pair(\Xx) = \left\{ y_i \,|\, (x_i,y_i) \in \Cc_\cf (\Xx, \Yy) \right\}
    \label{eq:ypair}
\end{equation}
and a residual set of samples that are never used for pairing,
\begin{equation}
    \Yy_\res(\Xx) = \Yy \backslash \Yy_\pair(\Xx)\,.
\end{equation} 
Intuitively, we expect that the average performance of the paired subset should be closer to the one of the reference set, $\mu[P(\Xx)]$, while the residual set shifts in the opposite direction.\footnote{As $\Yy_\res(\Xx) \cap \Yy_\pair(\Xx) = \emptyset$ we have $\mu[P(\Yy)] = \allowbreak \left( \mu[P(\Yy_\res(\Xx))]\, |\Yy_\res(\Xx)|\allowbreak + \mu[P(\Yy_\pair(\Xx))]\, |\Yy_\pair(\Xx)|\right)/|\Yy|$}

We demonstrate this approach by contrasting the weakly performing asset 9 with the four highest-performing assets.
In the three cases where those asset sets contain more samples, we use the samples from asset 9 to split those sets into paired and residual sets, respectively.
As seen from Table~\ref{tab:residuals} the paired sets have a performance that is comparable to the one of 9.
The other way around, the residual sets are performing better than the slices as a whole.
Using the smaller but equally well-performing set of asset 8, we can also split the slice of asset 9 into two, of which the paired one shows high performance.
Importantly, although these sub-divisions have a strong impact on the observed performance, they are determined purely based on the semantic properties of the elements.
Having ruled out \assid\ as a likely cause, the residual sets provide a reduced dataset for further, iterative exploration of the underlying semantic cause of the observable weakness.

\begin{table}[!htb] \caption{\label{tab:residuals} The average performance of residual and nearest neighbor subsets of $\Yy$.}
\centering
\begin{tabular}{ccc}
 Asset slices $(\Xx,\Yy)$   & $\mu[P(\Yy_\res(\Xx))]$  & $\mu[P(\Yy_\pair(\Xx))]$   \\
 \hline
  9, 4  & 0.59947 &  0.41172 \\  
  9, 24 & 0.60399 & 0.42597 \\ 
  9, 26 & 0.61561 & 0.37700 \\ 
  8, 9  & 0.39085 & 0.50945  \\ 
  \hline
\end{tabular}
\end{table}

\section{Conclusion}
\label{section:conclusion}

In this work, we have motivated a more detailed investigation of performance differences to identify systematic weaknesses in DNNs.
This analysis is based on slicing, where semantic features of the data are used to form meaningful subsets.
Within our approach, these features stem from the data generation process but, in theory, can also be be obtained through other methods.
It is, however, impossible to attribute all potential performance influencing dimensions, which forms a limitation of many approaches that aim to find weak slices as non-annotated or non-discovered characteristics can contain unresolved weaknesses.
Nonetheless, even with a limited amount of information, it is computationally challenging to identify semantic weaknesses correctly as accounting for the interaction between them leads to a ``combinatorial explosion.''
However, it is also not straightforward to simply attribute performance loss to specific semantic features when studying them in isolation.
This is caused, for instance, by inhomogeneities and sparseness of the data in high dimensions.
For this reason, we propose a method inspired by counterfactual explanations where we identify neighboring points between two slices of data based on their semantic similarity.
As these points are neighbors in the $(n-1)$-dimensional feature space, i.e., all features except the semantic-feature-under-test that defines the respective slices, the influence of these additional features is reduced. 
This allows us to study whether the semantic-feature-under-test constitutes an actual weakness of the model while avoiding prohibitive computational costs of considering the interactions of all features.
In our experiment from the autonomous driving domain, we could thus show that when considering the pedestrian asset type as semantic-dimension-under-test, from the two weakest performing assets, only in one case, the asset membership is the likely factor for the degraded performance.
Such insight is valuable for further improvement of the DNN, as the generation of additional (training) data to mitigate weaknesses becomes costly for complex tasks such as object detection or semantic segmentation.
In a second experiment, as an extension to the one before, we investigate one asset in further detail, where the asset itself was not the cause of the performance degradation.
Here, we demonstrate that counterfactual matching can be used to sub-partition existing slices based on their semantic features such that the weakly performing subset is carved out.
This allows a more refined analysis and can help find the actually relevant semantic dimensions among the remaining $n-1$ features more easily, given that the sub-partition forms a smaller dataset.

\section{Acknowledgments}
This work has been funded by the German Federal Ministry for Economic Affairs and Climate Action as part of the safe.trAIn project.

\bibliography{aaai23}

\begin{thebibliography}{43}
\providecommand{\natexlab}[1]{#1}

\bibitem[{Akhtar and Mian(2018)}]{akhtar2018threat}
Akhtar, N.; and Mian, A. 2018.
\newblock Threat of adversarial attacks on deep learning in computer vision: A
  survey.
\newblock \emph{IEEE Access}, 6: 14410--14430.

\bibitem[{Atzmueller(2015)}]{atzmueller2015subgroup}
Atzmueller, M. 2015.
\newblock Subgroup discovery.
\newblock \emph{Wiley Interdisciplinary Reviews: Data Mining and Knowledge
  Discovery}, 5(1): 35--49.

\bibitem[{Buolamwini and Gebru(2018)}]{buolamwini2018gender}
Buolamwini, J.; and Gebru, T. 2018.
\newblock Gender shades: Intersectional accuracy disparities in commercial
  gender classification.
\newblock In \emph{Conference on fairness, accountability and transparency},
  77--91. PMLR.

\bibitem[{Chakraborty et~al.(2018)Chakraborty, Alam, Dey, Chattopadhyay, and
  Mukhopadhyay}]{chakraborty2018adversarial}
Chakraborty, A.; Alam, M.; Dey, V.; Chattopadhyay, A.; and Mukhopadhyay, D.
  2018.
\newblock Adversarial attacks and defences: A survey.
\newblock \emph{arXiv preprint arXiv:1810.00069}.

\bibitem[{Chen et~al.(2017)Chen, Papandreou, Kokkinos, Murphy, and
  Yuille}]{chen2017deeplab}
Chen, L.-C.; Papandreou, G.; Kokkinos, I.; Murphy, K.; and Yuille, A.~L. 2017.
\newblock Deeplab: Semantic Image Segmentation with Deep Convolutional Nets,
  Atrous Convolution, and Fully Connected CRFS.
\newblock \emph{IEEE Transactions on Pattern Analysis and Machine
  Intelligence}, 40(4): 834--848.

\bibitem[{Chung et~al.(2019)Chung, Kraska, Polyzotis, Tae, and
  Whang}]{chung2019slice}
Chung, Y.; Kraska, T.; Polyzotis, N.; Tae, K.~H.; and Whang, S.~E. 2019.
\newblock Slice finder: Automated data slicing for model validation.
\newblock In \emph{2019 IEEE 35th International Conference on Data Engineering
  (ICDE)}, 1550--1553. IEEE.

\bibitem[{Cordts et~al.(2016)Cordts, Omran, Ramos, Rehfeld, Enzweiler,
  Benenson, Franke, Roth, and Schiele}]{cordts2016cityscapes}
Cordts, M.; Omran, M.; Ramos, S.; Rehfeld, T.; Enzweiler, M.; Benenson, R.;
  Franke, U.; Roth, S.; and Schiele, B. 2016.
\newblock The Cityscapes Dataset for Semantic Urban Scene Understanding.
\newblock In \emph{Proceedings of the IEEE/CVF Conference on Computer Vision
  and Pattern Recognition}, 3213--3223.

\bibitem[{Dandl et~al.(2020)Dandl, Molnar, Binder, and Bischl}]{dandl2020multi}
Dandl, S.; Molnar, C.; Binder, M.; and Bischl, B. 2020.
\newblock Multi-objective counterfactual explanations.
\newblock In \emph{International Conference on Parallel Problem Solving from
  Nature}, 448--469. Springer.

\bibitem[{De~Vries et~al.(2019)De~Vries, Misra, Wang, and Van~der
  Maaten}]{de2019does}
De~Vries, T.; Misra, I.; Wang, C.; and Van~der Maaten, L. 2019.
\newblock Does object recognition work for everyone?
\newblock In \emph{Proceedings of the IEEE/CVF Conference on Computer Vision
  and Pattern Recognition Workshops}, 52--59.

\bibitem[{d'Eon et~al.(2022)d'Eon, d'Eon, Wright, and
  Leyton-Brown}]{d2022spotlight}
d'Eon, G.; d'Eon, J.; Wright, J.~R.; and Leyton-Brown, K. 2022.
\newblock The spotlight: A general method for discovering systematic errors in
  deep learning models.
\newblock In \emph{2022 ACM Conference on Fairness, Accountability, and
  Transparency}, 1962--1981.

\bibitem[{Dhurandhar et~al.(2018)Dhurandhar, Chen, Luss, Tu, Ting, Shanmugam,
  and Das}]{dhurandhar2018explanations}
Dhurandhar, A.; Chen, P.-Y.; Luss, R.; Tu, C.-C.; Ting, P.; Shanmugam, K.; and
  Das, P. 2018.
\newblock Explanations based on the missing: Towards contrastive explanations
  with pertinent negatives.
\newblock \emph{Advances in neural information processing systems}, 31.

\bibitem[{Dosovitskiy et~al.(2017)Dosovitskiy, Ros, Codevilla, Lopez, and
  Koltun}]{dosovitskiy2017carla}
Dosovitskiy, A.; Ros, G.; Codevilla, F.; Lopez, A.; and Koltun, V. 2017.
\newblock CARLA: An Open Urban Driving Simulator.
\newblock In \emph{Proceedings of the Conference on Robot Learning}, 1--16.
  PMLR.

\bibitem[{Eyuboglu et~al.(2022)Eyuboglu, Varma, Saab, Delbrouck, Lee-Messer,
  Dunnmon, Zou, and R{\'e}}]{eyuboglu2022domino}
Eyuboglu, S.; Varma, M.; Saab, K.; Delbrouck, J.-B.; Lee-Messer, C.; Dunnmon,
  J.; Zou, J.; and R{\'e}, C. 2022.
\newblock Domino: Discovering systematic errors with cross-modal embeddings.
\newblock \emph{arXiv preprint arXiv:2203.14960}.

\bibitem[{Fingscheidt, Gottschalk, and Houben(2022)}]{fingscheidt2022deep}
Fingscheidt, T.; Gottschalk, H.; and Houben, S. 2022.
\newblock Deep Neural Networks and Data for Automated Driving: Robustness,
  Uncertainty Quantification, and Insights Towards Safety.

\bibitem[{Gannamaneni, Houben, and Akila(2021)}]{Gannamaneni_2021_ICCV}
Gannamaneni, S.; Houben, S.; and Akila, M. 2021.
\newblock Semantic Concept Testing in Autonomous Driving by Extraction of
  Object-Level Annotations From CARLA.
\newblock In \emph{Proceedings of the IEEE/CVF International Conference on
  Computer Vision (ICCV) Workshops}, 1006--1014.

\bibitem[{Goyal et~al.(2019)Goyal, Wu, Ernst, Batra, Parikh, and
  Lee}]{goyal2019counterfactual}
Goyal, Y.; Wu, Z.; Ernst, J.; Batra, D.; Parikh, D.; and Lee, S. 2019.
\newblock Counterfactual visual explanations.
\newblock In \emph{International Conference on Machine Learning}, 2376--2384.
  PMLR.

\bibitem[{Guidotti et~al.(2018)Guidotti, Monreale, Ruggieri, Pedreschi, Turini,
  and Giannotti}]{guidotti2018local}
Guidotti, R.; Monreale, A.; Ruggieri, S.; Pedreschi, D.; Turini, F.; and
  Giannotti, F. 2018.
\newblock Local rule-based explanations of black box decision systems.
\newblock \emph{arXiv preprint arXiv:1805.10820}.

\bibitem[{Hendrycks et~al.(2019)Hendrycks, Mu, Cubuk, Zoph, Gilmer, and
  Lakshminarayanan}]{hendrycks2019augmix}
Hendrycks, D.; Mu, N.; Cubuk, E.~D.; Zoph, B.; Gilmer, J.; and
  Lakshminarayanan, B. 2019.
\newblock Augmix: A simple data processing method to improve robustness and
  uncertainty.
\newblock \emph{arXiv preprint arXiv:1912.02781}.

\bibitem[{Herrera et~al.(2011)Herrera, Carmona, Gonz{\'a}lez, and
  Del~Jesus}]{herrera2011overview}
Herrera, F.; Carmona, C.~J.; Gonz{\'a}lez, P.; and Del~Jesus, M.~J. 2011.
\newblock An overview on subgroup discovery: foundations and applications.
\newblock \emph{Knowledge and information systems}, 29(3): 495--525.

\bibitem[{Hesamian et~al.(2019)Hesamian, Jia, He, and
  Kennedy}]{hesamian2019deep}
Hesamian, M.~H.; Jia, W.; He, X.; and Kennedy, P. 2019.
\newblock Deep learning techniques for medical image segmentation: achievements
  and challenges.
\newblock \emph{Journal of digital imaging}, 32(4): 582--596.

\bibitem[{{High-Level Expert Group on AI (AI HLEG)}(2019)}]{hleg}
{High-Level Expert Group on AI (AI HLEG)}. 2019.
\newblock Ethics Guidelines for Trustworthy AI.
\newblock Technical report, European Commission.

\bibitem[{Joshi et~al.(2019)Joshi, Koyejo, Vijitbenjaronk, Kim, and
  Ghosh}]{joshi2019towards}
Joshi, S.; Koyejo, O.; Vijitbenjaronk, W.; Kim, B.; and Ghosh, J. 2019.
\newblock Towards realistic individual recourse and actionable explanations in
  black-box decision making systems.
\newblock \emph{arXiv preprint arXiv:1907.09615}.

\bibitem[{Kanamori et~al.(2020)Kanamori, Takagi, Kobayashi, and
  Arimura}]{kanamori2020dace}
Kanamori, K.; Takagi, T.; Kobayashi, K.; and Arimura, H. 2020.
\newblock DACE: Distribution-Aware Counterfactual Explanation by Mixed-Integer
  Linear Optimization.
\newblock In \emph{IJCAI}, 2855--2862.

\bibitem[{Karimi et~al.(2020)Karimi, Barthe, Balle, and
  Valera}]{karimi2020model}
Karimi, A.-H.; Barthe, G.; Balle, B.; and Valera, I. 2020.
\newblock Model-agnostic counterfactual explanations for consequential
  decisions.
\newblock In \emph{International Conference on Artificial Intelligence and
  Statistics}, 895--905. PMLR.

\bibitem[{Karimi, Sch{\"o}lkopf, and Valera(2021)}]{karimi2021algorithmic}
Karimi, A.-H.; Sch{\"o}lkopf, B.; and Valera, I. 2021.
\newblock Algorithmic recourse: from counterfactual explanations to
  interventions.
\newblock In \emph{Proceedings of the 2021 ACM conference on fairness,
  accountability, and transparency}, 353--362.

\bibitem[{LeCun, Cortes, and Burges(2010)}]{lecun2010mnist}
LeCun, Y.; Cortes, C.; and Burges, C. 2010.
\newblock MNIST handwritten digit database.

\bibitem[{Loh et~al.(2022)Loh, Hauschke, Puntschuh, and
  Hallensleben}]{VDE_SPEC_90012}
Loh, W.; Hauschke, A.; Puntschuh, M.; and Hallensleben, S. 2022.
\newblock VDE SPEC 90012 V1.0 - VCIO based description of systems for AI
  trustworthiness characterisation.
\newblock Technical report, Verband der Elektrotechnik Elektronik
  Informationstechnik e.V. (VDE).

\bibitem[{Lundberg and Lee(2017)}]{lundberg2017unified}
Lundberg, S.~M.; and Lee, S.-I. 2017.
\newblock A unified approach to interpreting model predictions.
\newblock \emph{Advances in neural information processing systems}, 30.

\bibitem[{Lyssenko et~al.(2021)Lyssenko, Gladisch, Heinzemann, Woehrle, and
  Triebel}]{lyssenko2021evaluation}
Lyssenko, M.; Gladisch, C.; Heinzemann, C.; Woehrle, M.; and Triebel, R. 2021.
\newblock From evaluation to verification: Towards task-oriented relevance
  metrics for pedestrian detection in safety-critical domains.
\newblock In \emph{Proceedings of the IEEE/CVF Conference on Computer Vision
  and Pattern Recognition}, 38--45.

\bibitem[{Mahajan, Tan, and Sharma(2019)}]{mahajan2019preserving}
Mahajan, D.; Tan, C.; and Sharma, A. 2019.
\newblock Preserving causal constraints in counterfactual explanations for
  machine learning classifiers.
\newblock \emph{arXiv preprint arXiv:1912.03277}.

\bibitem[{Mothilal, Sharma, and Tan(2020)}]{mothilal2020explaining}
Mothilal, R.~K.; Sharma, A.; and Tan, C. 2020.
\newblock Explaining machine learning classifiers through diverse
  counterfactual explanations.
\newblock In \emph{Proceedings of the 2020 conference on fairness,
  accountability, and transparency}, 607--617.

\bibitem[{Oza et~al.(2021)Oza, Sindagi, VS, and Patel}]{oza2021unsupervised}
Oza, P.; Sindagi, V.~A.; VS, V.; and Patel, V.~M. 2021.
\newblock Unsupervised domain adaptation of object detectors: A survey.
\newblock \emph{arXiv preprint arXiv:2105.13502}.

\bibitem[{Pearl(2019)}]{pearl2019seven}
Pearl, J. 2019.
\newblock The seven tools of causal inference, with reflections on machine
  learning.
\newblock \emph{Communications of the ACM}, 62(3): 54--60.

\bibitem[{Radford et~al.(2021)Radford, Kim, Hallacy, Ramesh, Goh, Agarwal,
  Sastry, Askell, Mishkin, Clark et~al.}]{radford2021learning}
Radford, A.; Kim, J.~W.; Hallacy, C.; Ramesh, A.; Goh, G.; Agarwal, S.; Sastry,
  G.; Askell, A.; Mishkin, P.; Clark, J.; et~al. 2021.
\newblock Learning transferable visual models from natural language
  supervision.
\newblock In \emph{International Conference on Machine Learning}, 8748--8763.
  PMLR.

\bibitem[{Ribeiro, Singh, and Guestrin(2016)}]{ribeiro2016should}
Ribeiro, M.~T.; Singh, S.; and Guestrin, C. 2016.
\newblock " Why should i trust you?" Explaining the predictions of any
  classifier.
\newblock In \emph{Proceedings of the 22nd ACM SIGKDD international conference
  on knowledge discovery and data mining}, 1135--1144.

\bibitem[{Sagadeeva and Boehm(2021)}]{sagadeeva2021sliceline}
Sagadeeva, S.; and Boehm, M. 2021.
\newblock Sliceline: Fast, linear-algebra-based slice finding for ml model
  debugging.
\newblock In \emph{Proceedings of the 2021 International Conference on
  Management of Data}, 2290--2299.

\bibitem[{Siam et~al.(2018)Siam, Gamal, Abdel-Razek, Yogamani, Jagersand, and
  Zhang}]{siam2018comparative}
Siam, M.; Gamal, M.; Abdel-Razek, M.; Yogamani, S.; Jagersand, M.; and Zhang,
  H. 2018.
\newblock A comparative study of real-time semantic segmentation for autonomous
  driving.
\newblock In \emph{Proceedings of the IEEE conference on computer vision and
  pattern recognition workshops}, 587--597.

\bibitem[{Syed~Sha, Grau, and Hagn(2020)}]{syed2020dnn}
Syed~Sha, Q.; Grau, O.; and Hagn, K. 2020.
\newblock DNN analysis through synthetic data variation.
\newblock In \emph{Computer Science in Cars Symposium}, 1--10.

\bibitem[{Ustun, Spangher, and Liu(2019)}]{ustun2019actionable}
Ustun, B.; Spangher, A.; and Liu, Y. 2019.
\newblock Actionable recourse in linear classification.
\newblock In \emph{Proceedings of the conference on fairness, accountability,
  and transparency}, 10--19.

\bibitem[{Verma, Dickerson, and Hines(2020)}]{verma2020counterfactual}
Verma, S.; Dickerson, J.; and Hines, K. 2020.
\newblock Counterfactual explanations for machine learning: A review.
\newblock \emph{arXiv preprint arXiv:2010.10596}.

\bibitem[{Wachter, Mittelstadt, and Russell(2017)}]{wachter2017counterfactual}
Wachter, S.; Mittelstadt, B.; and Russell, C. 2017.
\newblock Counterfactual explanations without opening the black box: Automated
  decisions and the GDPR.
\newblock \emph{Harv. JL \& Tech.}, 31: 841.

\bibitem[{Wang and Deng(2018)}]{wang2018deep}
Wang, M.; and Deng, W. 2018.
\newblock Deep visual domain adaptation: A survey.
\newblock \emph{Neurocomputing}, 312: 135--153.

\bibitem[{Wang et~al.(2020)Wang, Qinami, Karakozis, Genova, Nair, Hata, and
  Russakovsky}]{wang2020towards}
Wang, Z.; Qinami, K.; Karakozis, I.~C.; Genova, K.; Nair, P.; Hata, K.; and
  Russakovsky, O. 2020.
\newblock Towards fairness in visual recognition: Effective strategies for bias
  mitigation.
\newblock In \emph{Proceedings of the IEEE/CVF conference on computer vision
  and pattern recognition}, 8919--8928.

\end{thebibliography}

\end{document}